\documentclass[10pt,twocolumn,letterpaper]{article}

\usepackage{iccv}
\usepackage{times}
\usepackage{epsfig}
\usepackage{graphicx}
\usepackage{amsmath}
\usepackage{amssymb}

\usepackage{booktabs}
\usepackage{multirow}
\usepackage{subfig}
\usepackage{tikz}
\usetikzlibrary{tikzmark}
\usepackage{bm}

\makeatletter
\renewcommand\paragraph{\@startsection{paragraph}{4}{\z@}%
                                    {2.3ex \@plus1ex \@minus.2ex}%
                                    {-1em}%
                                    {\normalfont\normalsize\bfseries}}
\makeatother

\usepackage[pagebackref=true,breaklinks=true,letterpaper=true,colorlinks,bookmarks=false]{hyperref}

\iccvfinalcopy 

\def\iccvPaperID{3752} 

\ificcvfinal\fi
\begin{document}

\title{Spatial Shortcut Network for Human Pose Estimation}

\author{Te Qi\qquad Bayram Bayramli\qquad Usman Ali\qquad Qinchuan Zhang\qquad Hongtao Lu\thanks{Corresponding author.} \\
Shanghai Jiao Tong University\\
{\tt\small \{qite1030,bayram\_bai,usmanali,qinchuan.zhang,htlu\}@sjtu.edu.cn}
}

\maketitle

\begin{abstract}

  Like many computer vision problems, human pose estimation is a challenging problem in that recognizing a body part requires not only information from local area but also from areas with large spatial distance. In order to spatially pass information, large convolutional kernels and deep layers have been normally used, introducing high computation cost and large parameter space. Luckily for pose estimation, human body is geometrically structured in images, enabling modeling of spatial dependency. In this paper, we propose a spatial shortcut network for pose estimation task, where information is easier to flow spatially. We evaluate our model with detailed analyses and present its outstanding performance with smaller structure. The code will be published after paper being accepted.
\end{abstract}


\section{Introduction}

\begin{figure}[t]
  \centering
    \includegraphics[width=\linewidth]{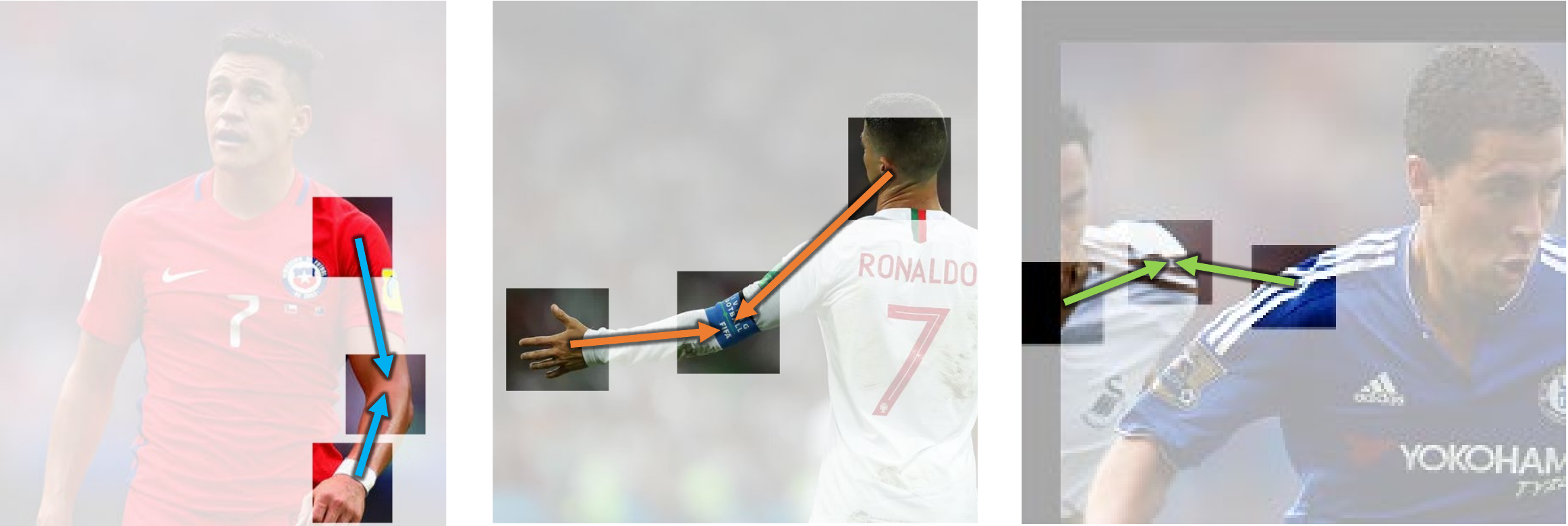}
  \caption{Examples of long-range spatial dependency. Arrows indicate dependency paths. \textbf{Left}: Hints from shoulders and wrists for elbow detection. \textbf{Middle}: Left-right side discrimination depends on head and hand orientation. \textbf{Right}: Left shoulder of the non-primary person can be suppressed if nearby image border and another parts are both ``seen''.}
  \label{fig:motivation}
\end{figure}

Human pose estimation is a problem with strong long-range spatial dependency. Though in many methods the task is formed as a localized per-pixel classifying problem, the feature extraction process is still non-local. The existence of a body part can not be simply determined with only local features, but also information from surrounding areas or even other body parts. We show three examples in Figure~\ref{fig:motivation} to demonstrate the importance of long-range dependency. In the left image, since the appearance of an elbow is very similar to that of a knee, it is not easy to discriminate the two for a small feature extractor whose receptive field can only cover the elbow itself. But if the extractor can also ``see'' the nearby wrist or shoulder in the same time, the classifying of it being elbow can be much easier. Similarly in the middle image, to determine whether a body part is a left or right one, the orientation of the person's head and hand is significant information. In methods involving single-person pose estimation, the detection of non-primary person's body parts need to be suppressed. As in the right image, with the information from nearby person and image border, feature extractors can suppress the shoulder's detection.

Our work intends to improve the modeling of spatial dependency in CNN-based pose estimation methods. This allows more efficient and quicker pose summarizing process, thus our model can achieve better performance with smaller structure.

In traditional methods, pictorial structures with handcrafted feature detectors are used to extract body parts. Pictorial structures assume priors on distance and angle distribution between parts, explicitly modeling spatial feature relation. With the success of deep neural networks, the long-range dependency can be modeled by kernel parameters implicitly. Many methods use very deep networks to allow more global pose summarizing. In these methods, large convolutional kernels, pooling and deep stack of layers are generally three strategies for promoting spatially information flowing. Large kernel provides broad receptive field, but its computation is expensive and the learned filters are less generalized. Besides, with limited computation resources, the chosen kernel size still can not be large enough. Alternatively, very deep networks and pooling layers can learn the dependency more generally. But this inevitably means many layers for a small distance, thus gradient vanishing problem can hinder the dependency's training.

\begin{figure*}[t]
  \centering
  \includegraphics[width=0.33\linewidth]{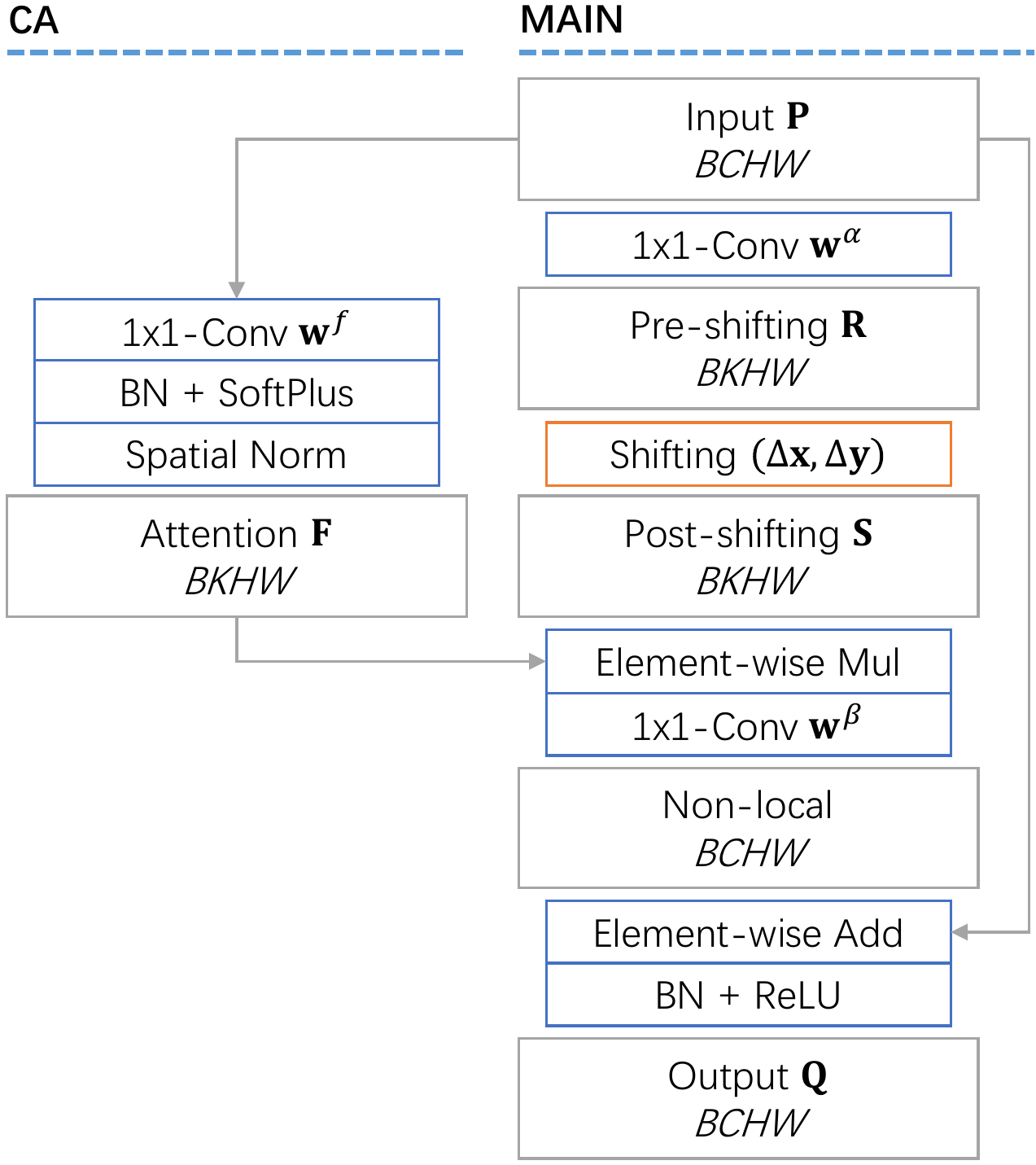}\hspace{0.02\linewidth}
  \includegraphics[width=0.58\linewidth]{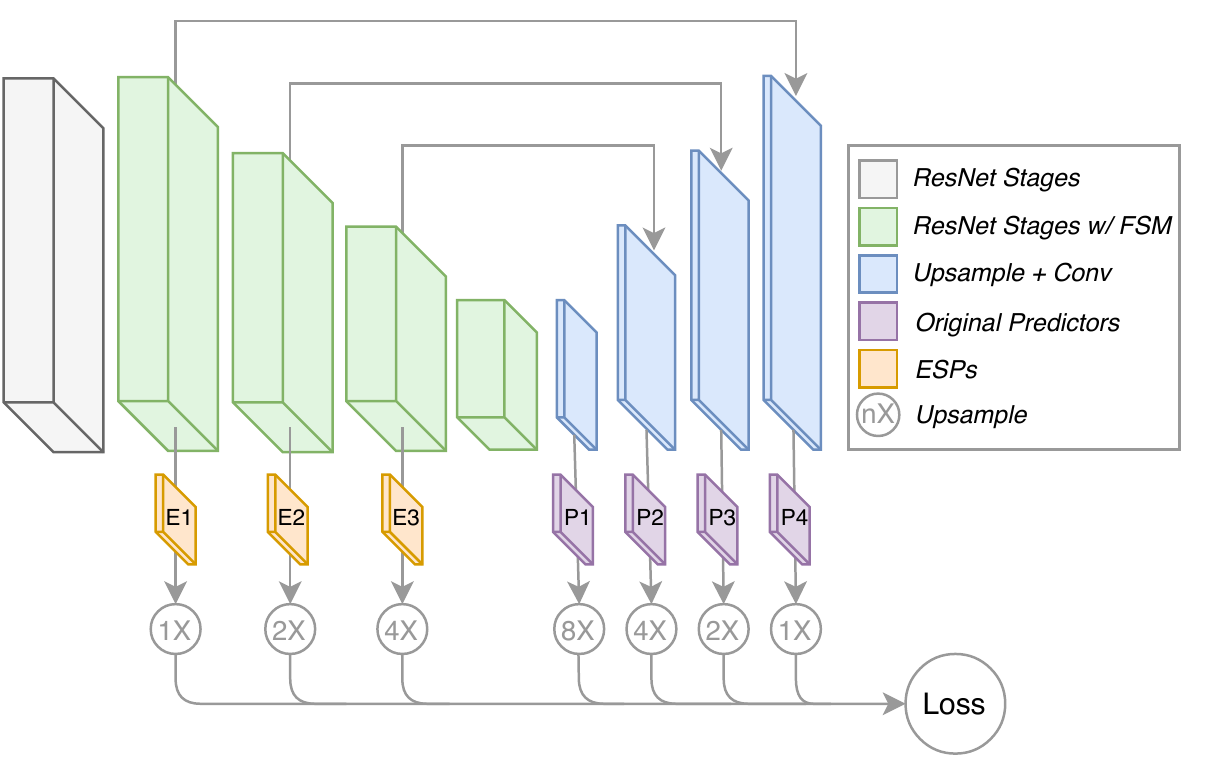}
  \caption{Structure of proposed model. \textbf{Left}: Feature shifting module. Wide rectangles are feature maps, narrow ones are operations. \textit{BCHW} or \textit{BKHW} represents the map's shape in the order: batch size, channel number, height and width. \textbf{Right}: Spatial shortcut network. Original predictors are \textit{P*} and early stage predictors are \textit{E*}. }
  \label{fig:fsm_n_network}
\end{figure*}

To improve the modeling of long-range dependency, we propose a network called spatial shortcut network (SSN), where features are especially easier to flow spatially. In our method, we combine feature map shifting and attention mechanism in a module called feature shifting module (FSM). This module establishes spatial shortcut tunnels for features to pass through, with the tunnels' offsets optimizable. Its decoupling of channels and attention mechanism make it equivalent to an efficient convolution layer with very large and input-dependent convolution window. The structure of FSM and SSN are shown in Figure~\ref{fig:fsm_n_network}. In this paper, our contributions are summarized as follows
\begin{itemize}
  \item We propose the feature shifting module to learn to establish spatial shortcut tunnels. We demonstrate its feature map shifting, channel decoupling and attention mechanism forms it a window-optimizable, efficient and flexible convolution layer.
  \item The learned FSMs are analyzed in detail. Its ability in the modeling of spatial dependency and relation between keypoint detection and shifting offsets are shown.
  \item We show our model can achieve good or even better result with smaller structure. A very lightweight network with competitive performance is also presented, allowing applications on devices with limited resource.
\end{itemize}

\section{Related works}

\paragraph{Pose Estimation}


Traditional pose estimation methods are based on handcrafted features~\cite{dalal_cvpr_2005} and pictorial structures~\cite{andriluka_cvpr_2009}. With the development of convolutional neural networks, the feature extraction part of most methods are replaced with CNNs. Toshev and Szegedy~\cite{toshev_cvpr_2014} propose a multi-stage CNN-based regressor to regress and gradually refine poses from previous stage. Wei \etal~\cite{wei_cvpr_2016} also use a multi-stage model with interleaved large kernel convolution and pooling layers to achieve large receptive field. Hourglass Network by Newell \etal~\cite{newell_eccv_2016} replaces each stage by nested layers of pooling and up-sampling, and uses shortcut connections to capture multi-scale features. Fang \etal~\cite{fang_iccv_2017} use STN~\cite{jaderberg_nips_2015} to correct bounding box proposals from person detector before they are used to crop images.

Though prediction-based methods can preserve some extent of resolution for localization, they only have pixel level precision. In \cite{papandreou_cvpr_2017}, additional offset fields are regressed for refinement to achieve sub-pixel precision. CPN~\cite{chen_cvpr_2018} applies a RefineNet on predicted maps to actively refine on hard keypoints.

Bottom-up approaches are also proposed. DeepCut~\cite{pishchulin_cvpr_2016} and DeeperCut~\cite{insafutdinov_eccv_2016} both build a probability estimator on feasible solutions of pairwise keypoints. PAFs~\cite{cao_cvpr_2017} learns to predict fields of limb directions between every two adjacent keypoints. PPN~\cite{sekii_eccv_2018} forms detecting body part as an object detection problem, regressing limb offsets and parsing poses in a probabilistic manner.


\paragraph{Spatial Denpendency and Receptive Field}

Some pose estimation methods have also modeled spatial dependency. Chu \etal~\cite{chu_cvpr_2017} use conditional random fields~\cite{lafferty_icml_2001} and attention mechanism~\cite{bahdanau_iclr_2015,ba_iclr_2015} to exploit contextual information from surrounding area. Non-local network~\cite{wang_cvpr_2018} models contextual information by applying spatial correlation. However the learned correlation in non-local network is between absolute positions instead of relative ones, leading to insufficient generalization ability.

Variants or combinations of traditional convolution have also been designed. Atrous convolution~\cite{yu_iclr_2016} has strided convolutional kernels to achieve larger receptive field with no parameter increment. ASPP~\cite{chen_arxiv170605587cs_2017} adopts parallel atrous convolution with different atrous rates to capture multi-scale context. RFB~\cite{liu_eccv_2018l} further improved this by adding large convolution kernels of different sizes before each atrous convolution, which is a simulation of human population receptive field.

Some other methods adopt learnable receptive fields, which is also used in our model. ShiftNet~\cite{wu_cvpr_2018} learns integer offsets for shifting feature maps. Deformable CNN~\cite{dai_iccv_2017} regress their convolution kernel offset with fractional value for every channel and every spatial position. Similarly, active convolution~\cite{jeon_cvpr_2017} uses optimizable kernel offset without regressing, and the offset values are consistent across spatial positions. Our method can also be view as learning kernel offset, but it is more efficient than deformable convolution and active convolution, which we will introduce in Section~\ref{sec:fsm}. Similar to our method, Jeon and Kim~\cite{jeon_arxiv180607370cs_2018} use fractional learnable offset for shifting of each channel. But in our method the decoupling of channels and the introducing of attention makes it a supplementary part to other networks, with its learned offsets more dedicated to modeling long-range dependency.

\section{Approach}

The most important part of our method is feature shifting module. The module is as lightweight as a convolution layer in both parameter number and computation cost, and can be inserted to any part of a network for supplementing spatially summarized information. In this section, we will introduce the module firstly and then present in detail its attention mechanism, named as \textit{correlation attention} (CA). At last the whole framework will be introduced.

\subsection{Feature shifting module}\label{sec:fsm}

The structure of feature shifting module is shown in Figure~\ref{fig:fsm_n_network}'s left. It consists of two branches: the main branch and CA branch. In this section we focus on the main branch. This branch takes \(C\) channels of feature maps \(\mathbf{P}\) as input. The maps are transformed into \(K\) channels by the first \(1\times 1\)-conv. Then the shifting layer apply per-channel shifting operation with \(K\) pairs of offset parameters. The shifted maps are element-wise multiplied with attention maps that are generated by CA branch, and transformed back to \(C\) channels by another \(1\times 1\)-conv. Finally they are added with shortcut of input maps, with batch normalization and non-linear operation followed.


\begin{figure}[t]
  \centering
  \includegraphics[width=0.7\linewidth]{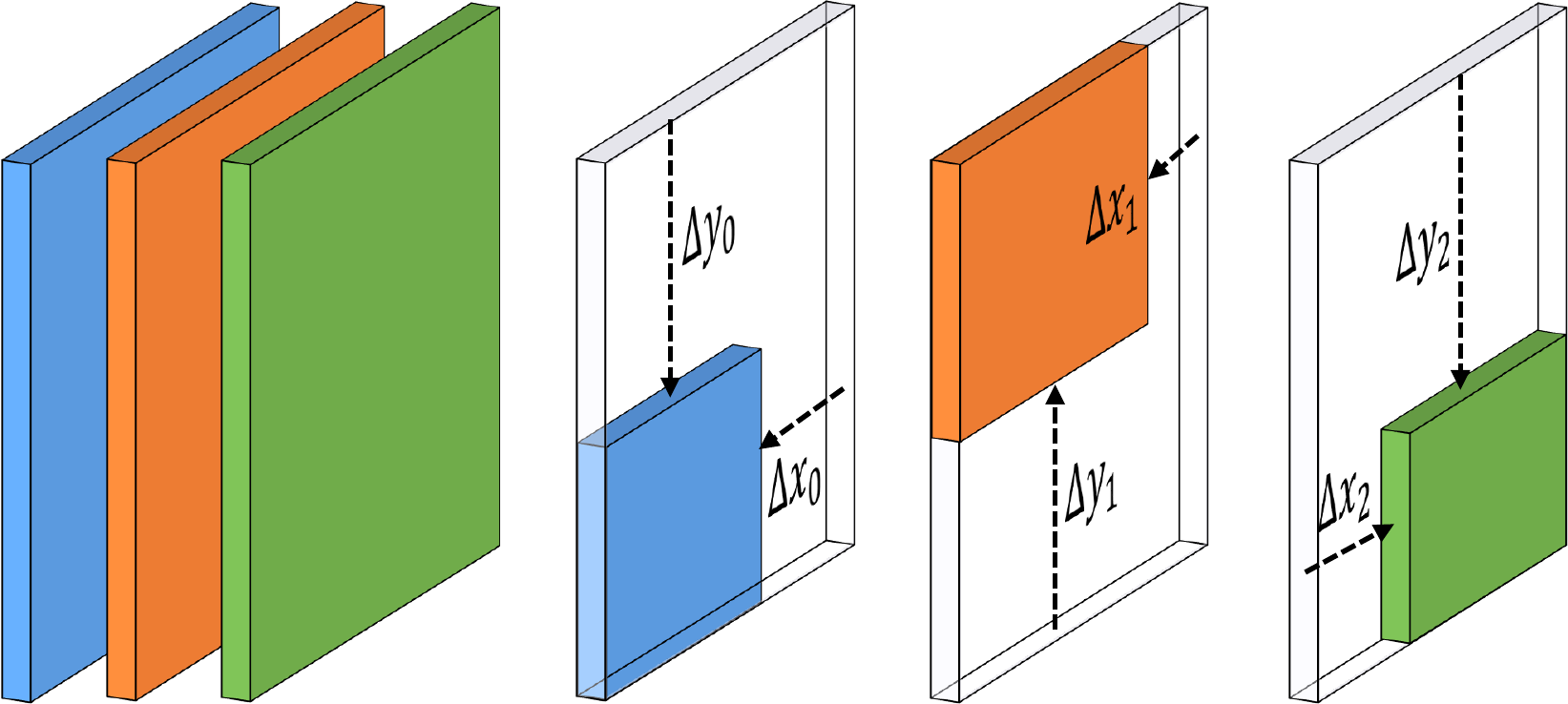}
  \caption{Shifting operation with \((\mathbf{\Delta x}, \mathbf{\Delta y})\) when \(K=3\).}
  \label{fig:shift}
\end{figure}

\begin{figure*}[t]
    \centering
    \includegraphics[width=\linewidth]{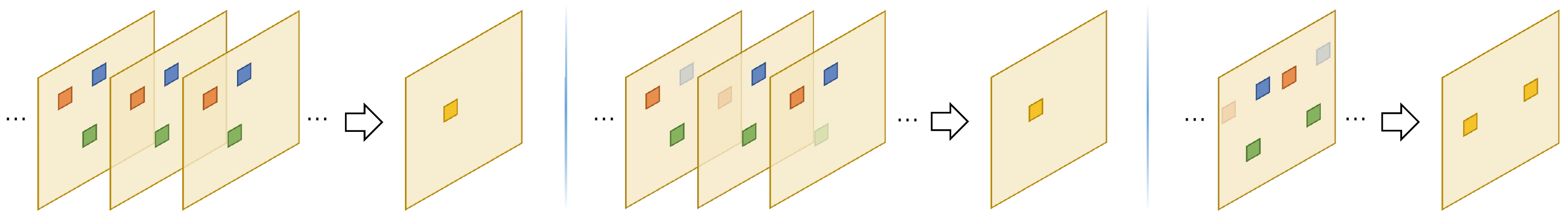}
    \caption{Illustration of convolutions over all \(C\) input channels when \(K\) = 3. \textbf{Left}: Active convolution and deformable convolution operate with all \(K\) window positions in each input channel. \textbf{Middle}: FSM operates with subset of window positions. \textbf{Right}: With CA, different window shapes can be formed within one input channel at different output positions, depending on input data at each position.}
    \label{fig:3_conv}
\end{figure*}

The shifting operation is illustrated in Figure~\ref{fig:shift}. With \(K\) channels of feature maps and \(K\) pairs of horizontal and vertical offsets \((\mathbf{\Delta x}, \mathbf{\Delta y}) \in \mathbb{R}^{K \times 2}\), this layer translates each map with the corresponding offset pair. The pixels translated out of view are ignored and the ones translated in are filled with zero. To make the offset parameters optimizable during training, real-number offsets are used and we adopt bilinear interpolation for the fraction part of the offsets. Denote the \(K\) pre-shifting feature maps by \(\mathbf{R}\) and post-shifting maps by \(\mathbf{S}\), the shifting operation can be formulated as
\begin{equation}
    S_k(x, y) = R_k^*(x-\Delta x_k, y-\Delta y_k), k=1, \dots, K
\end{equation}
where \(R_k^*\) is bilinear interpolated map of \(R_k\). Denote the parameters of the \(1\times 1\)-conv before shifting by \(\mathbf{w^{\alpha}}\), \(R_k\) is given by
\begin{equation}
    R_k(x, y) = \sum_{c=1}^C w^{\alpha}_{k,c} P_c(x, y).
\end{equation}

To help in illustrating the module's function, the interpolation on \(\mathbf{R}\) can also be interpreted to interpolation on \(\mathbf{P}\) since both bilinear interpolation and convolution are linear operation
\begin{equation}
    R_k^*(x, y) = \sum_{c=1}^C w^{\alpha}_{k,c} P^*_c(x, y).
\end{equation}

Therefore, the feature shifting module can be formulated as\footnote{We abbreviate \((x, y)\) to \((\cdot)\) here for simplicity.}
\begin{equation}
    \begin{aligned}
        &Q_c(\cdot) \\
        =& \sigma \left( P_c(\cdot) + \sum^K_{k=1} w^{\beta}_{c,k} F_k(\cdot) S_k(\cdot) \right) \\
        =& \sigma \left( P_c(\cdot) + \sum^K_{k=1} \sum_{c'=1}^C w^{\mathrm{FSM}}_{c,k,c'}(\cdot) P^*_{c'}(x-\Delta x_k, y-\Delta y_k) \right)
    \end{aligned}
\end{equation}
where \(Q_c(\cdot)\) is \(c\)-th output channel of FSM, \(w^{\beta}_{c,k}\) is the weight of the post-shifting \(1\times 1\)-conv, \(F_k(\cdot) \in (0, 1)\) is correlation attention, and
\begin{equation}\label{eq:fsm_weight}
    w^{\mathrm{FSM}}_{c,k,c'}(x, y) = w^{\beta}_{c,k} w^{\alpha}_{k,c'} F_k(x, y)
\end{equation}

It is now obvious that FSM can be viewed as approximation of a convolution layer plus shortcut connection and nonlinear function. The weights of this convolution are \(\mathbf{w}^{\mathrm{FSM}}\), which are constructed by the two \(1\times 1\)-conv's weights and modulated by attention. The window of this convolution is not in traditional grid shape, e.g. a \(3\times 3\) grid in \(3\times 3\)-conv, but in a shape defined by all \(K\) offsets. When under this convolution view, we refer to the convolution positions in input maps as \textit{window positions}, as is the orange/blue/green dots in Figure~\ref{fig:3_conv}.

\paragraph{Channel Decoupling}
The reason we add \(1\times 1\)-conv before and after shifting layer is to decouple backbone channels (input channels) and shifting channels. Firstly, if applying shifting directly on input channels, not every one of them needs to be shifted, and there would be no way to shift a channel by multiple different offsets. Secondly, the number of needed shifting channels does not solely depend on the number of input channels. The goal of shifting is to align spatially related positions together, thus the number of offsets needed should also depend on spatial distribution of features. Thirdly, we expect FSMs to provide supplemental information to backbone, and the backbone in our method is also pretrained without FSMs on ImageNet~\cite{russakovsky_int.j.comput.vis._2015}. With decoupled shifting channels, after FSMs are inserted we can prevent backbone's training from being radically disturbed, thus they can provide additive performance.

Compared to an active convolution~\cite{jeon_cvpr_2017} or deformable convolution~\cite{dai_iccv_2017} layer, FSM is more efficient in parameters. With the number of input and output channels both \(C\) and covering \(K\) window positions, FSM has totally \(3KC+2K\) parameters. For active convolution and deformable convolution, to cover \(K\) window positions as well, the numbers of parameters are \(KC^2 + 2K\) and \(KC^2 + 2KC\) respectively. The difference is that they convolute each input channel with all \(K\) window positions, as shown in Figure~\ref{fig:3_conv}'s left and middle. However not every window position is necessary in all channels, especially in a convolution layer with large \(K\). By decoupling input and shifting channels, FSM is equivalent to selecting within each input channel only a subset of window positions for convoluting with, and the subset can be different among channels. This is illustrated in the middle of Figure~\ref{fig:3_conv}. We believe FSM is more efficient because the features with long-range spatial dependency might seldom co-exist within single input channel. Taking the example of detecting elbows in the left of Figure~\ref{fig:motivation}, features for shoulders and wrists are likely in separated channels.

\subsection{Correlation attention}

We introduce correlation attention (CA) to regulate where and how FSM should be effective based on input data. Without it, FSM will indiscriminately convolute at every spatial position, producing noise at positions without spatial dependency or probably over-generalizing the learned dependency. Correlation attention predicts at every spatial position whether spatial dependency exists, or we can say whether shifted features are correlated with local features.

To predict the correlation confidence, ideally we should use both pre- and post-shifting maps as the source, but experiments show this has similar performance to just using input maps. Hence the CA branch is given by
\begin{equation}
  \begin{aligned}
    F_k(x, y) =& \frac{f_k(x,y)}{\sum_{(x',y') \in \Omega} f_k(x',y')} \\
    f_k(x, y) =& \sigma \left( \sum_{c=1}^C w^f_{k,c} P_c(x, y) \right)
  \end{aligned}
\end{equation}
where \(\Omega\) is the set of all spatial positions, and \(w^f_{k,c}\) is a weight of the \(1\times 1\)-conv in the CA branch.

From equation~\ref{eq:fsm_weight}, we can see CA is acting as a gate for each window position. This allows not only toggling of the whole convolution, but also partly activating, forming different window shape based on input data. We show this in the right of Figure~\ref{fig:3_conv} and will visualize it in the analyses section.

\subsection{Backbone network and early stage predictor}


\begin{figure*}[t]
    \centering
    \begin{minipage}[b]{0.37\linewidth}
      \subfloat[FSM]{\includegraphics[width=\linewidth]{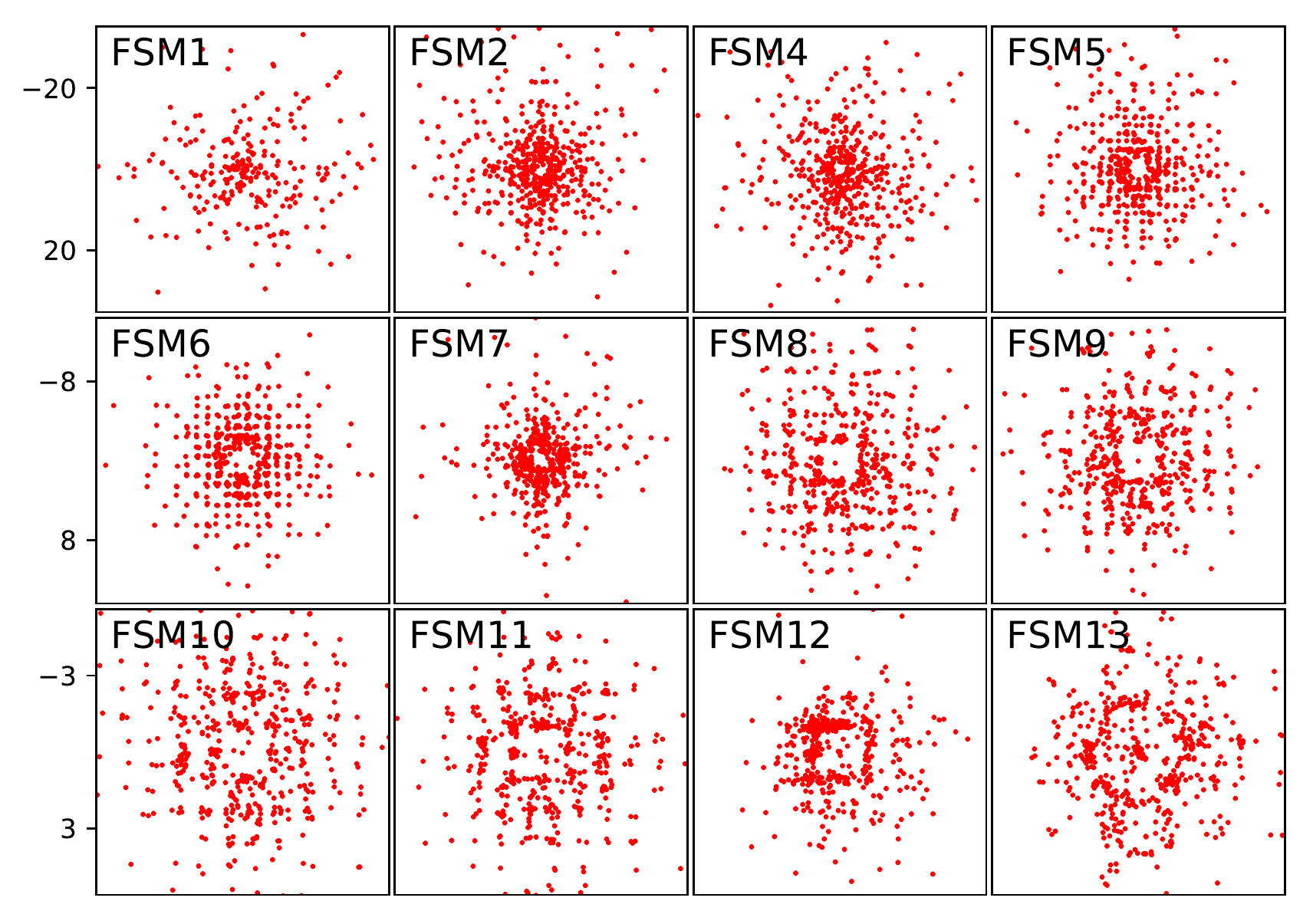}}\\
      \subfloat[FSM w/o CA]{\includegraphics[width=\linewidth]{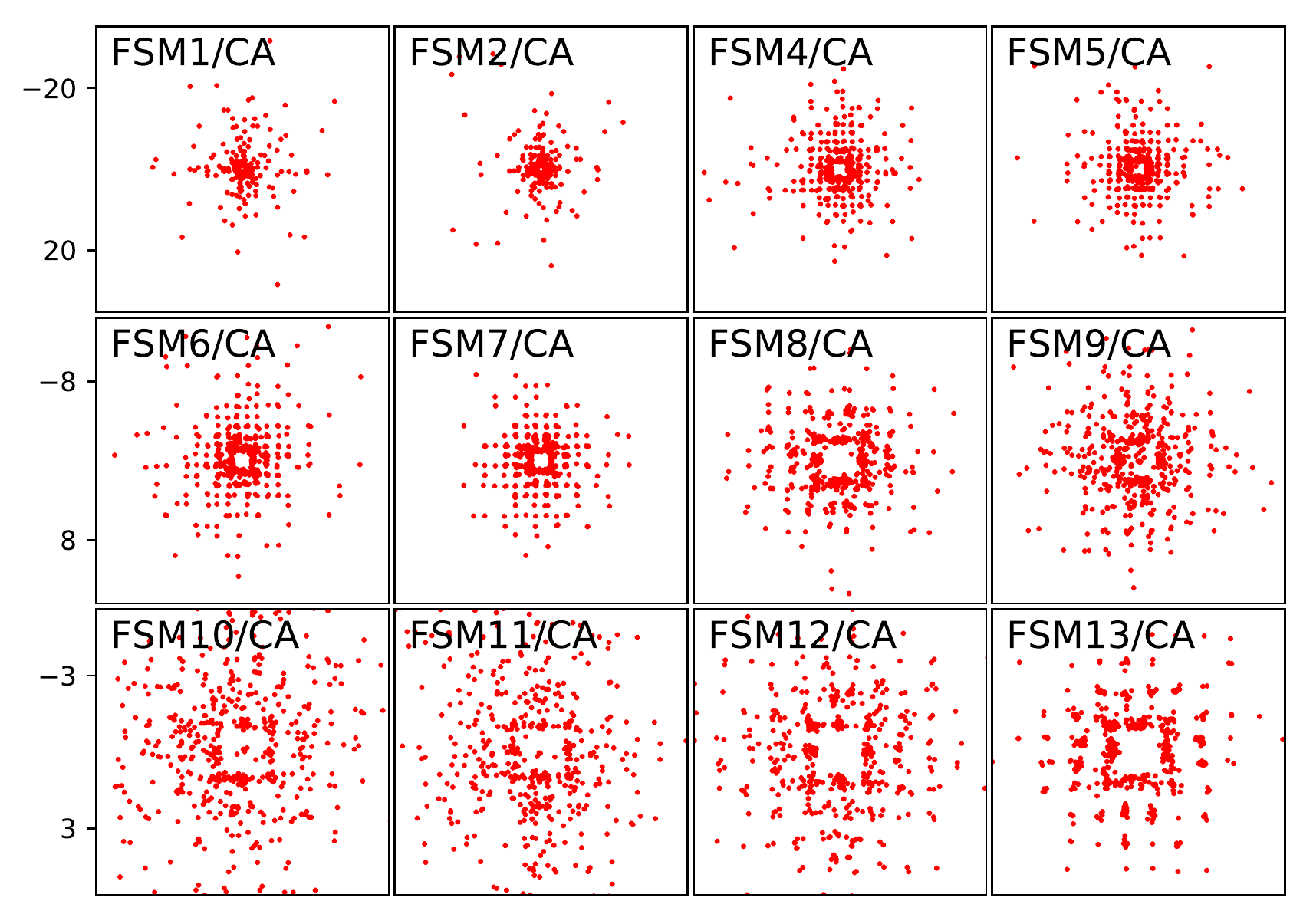}}
    \end{minipage}
    \subfloat[FSM3]{\includegraphics[width=0.31\linewidth]{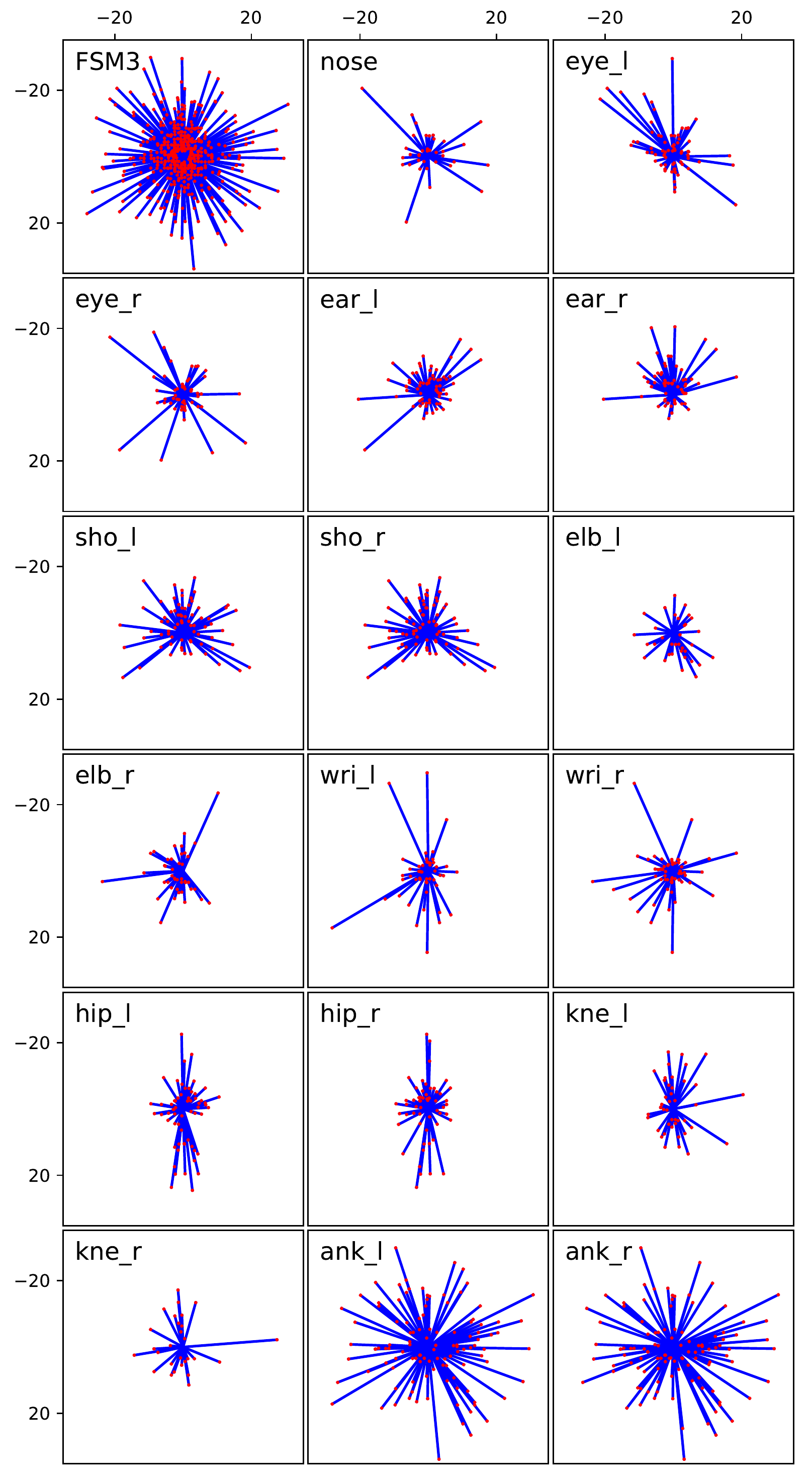}}
    \subfloat[FSM3 w/o CA]{\includegraphics[width=0.31\linewidth]{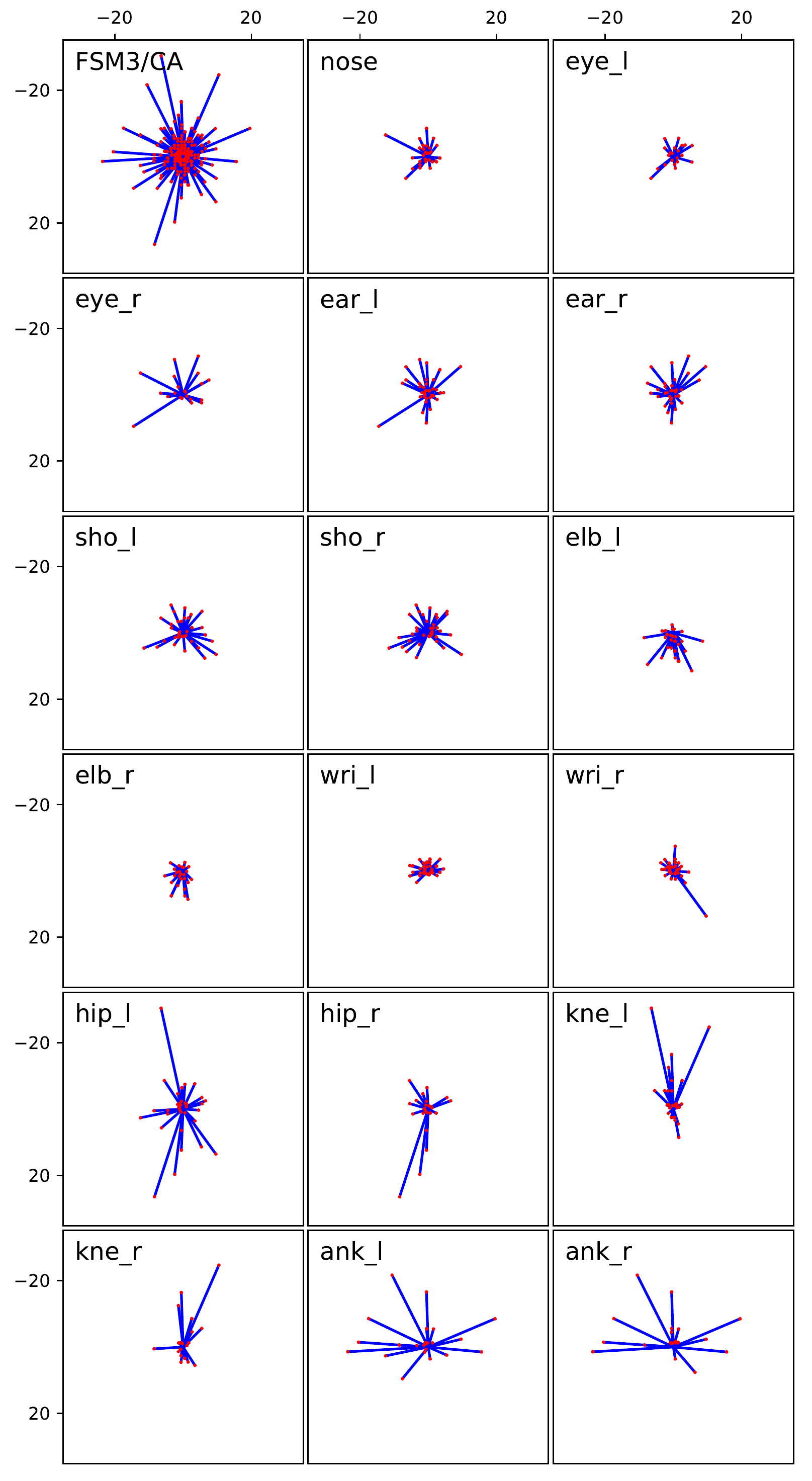}}
  \caption{Learned offsets of all FSMs and keypoint-related offsets of FSM3}\label{fig:all_offset}
\end{figure*}



The structure of our SSN is presented in Figure~\ref{fig:fsm_n_network}. We use a U-shaped network as our backbone, which is equal to CPN~\cite{chen_cvpr_2018} without its RefineNet. It is similar to a FPN~\cite{lin_cvpr_2017a} and is the combination of ResNet~\cite{he_cvpr_2016}, up-sampling layers and shortcut connections between them. 


In a deep CNN model, features from shallow layers can provide localization precision but are not good at classifying, while deep features are opposite. U-shaped networks combine their advantage by adding them together. We believe that the poor classifying ability of shallow layers is mainly due to their small receptive field. Though deeper layers can cover large area of an image, they are also limited in spatially summarizing from previous layer. Therefore we insert FSMs before every Bottleneck block in ResNet to widen their receptive field. With broad receptive field of every part, more pose summarizing and refinement can be allowed in the network.

FPN~\cite{lin_cvpr_2017a} proposed to add extra predictors on up-sampled feature maps. This applies intermediate supervision on its up-sampling layers, and also shallow layers due to its shortcut connections. To focus more on improving shallow layers whose detecting ability could be boosted by FSMs, we added more predictors directly on shallow layers, which we name as early stage predictors (ESP) and denote by \textit{E*} in Figure~\ref{fig:fsm_n_network}. In addition to intermediately supervising, with ESPs it is easier to see how the detecting ability evolves throughout the network, and how FSMs improve every part.

\section{Experiment}

\paragraph{Implementation Details}

To stabilize the training process with small batch size, we use group normalization \cite{wu_eccv_2018i} instead of batch normalization~\cite{ioffe_icml_2015} in the backbone. Group normalization shows consistent training ability over different batch sizes. For normalization inside the FSM, since FSMs' natural channel unbalance is conflict to the assumption of group normalization, we still use batch normalization in it.

There are 16 Bottleneck blocks in ResNet-50, with 3, 4, 6 and 3 blocks in four stages. FSMs are inserted before every block, except for places right after the pooling blocks. We have observed that FSMs tend to compensate for lost information during pooling, which induces shifting offsets falling into local minima. So the four stages have 3, 3, 5 and 2 FSMs respectively.

Unless specifically stated, the number of shifting channels \(K\) is set to 512 for all the FSMs except the first one, whose \(K\) is 256. 



\paragraph{Dataset and Data Augmentation}

We use Microsoft COCO dataset~\cite{lin_arxiv14050312cs_2014} to validate our model. For training, the \textit{train2017} subset, also called \textit{trainval35k}, is used. It includes 57k images and 150k person instances. The \textit{val2017} subset (a.k.a. \textit{minival}) contains 5k images for validation. We also test our model on \textit{test-dev} subset with 20k images. The performance of our model on COCO is reported in OKS-based mAP. OKS means object keypoint similarity, which is intuitively similar to IoU (intersection over union) used in COCO object detection.


We adopted data augmentation process during training. The images cropped from ground truth bounding boxes are randomly rotated by \(-30^{\circ} \sim 30^{\circ}\), scaled by factor of \(0.75 \sim 1.25\) and shifted by \(-0.05 \sim 0.05\) times of image size.

\paragraph{Training}

The backbone network is trained with Adam algorithm~\cite{kingma_iclr_2015} at learning rate of 5e-4. The batch size is 16. After 300K iterations, we degrade the learning rate by a factor of 2 for every 30k iterations and stop training after 400k iterations. ResNet without FSMs are firstly pretrained on ImageNet~\cite{russakovsky_int.j.comput.vis._2015}.

If FSMs are directly trained from the beginning, the shifting offsets will receive large gradient and thus move the whole feature map out of view, in which case they will never have a chance to move back. So a delayed insertion strategy is applied; FSMs are inserted after the other part of the network has already been trained for 6000 iterations. This is similar to the warming up strategy in \cite{jeon_cvpr_2017}.

A different learning rate and its degrading strategy for offsets are also adopted. At the beginning, their learning rate is set to 1e-3 for better searching. However changeful offsets will also harm the network's convergence. Therefore we decrease their learning rate by 10\% after each epoch.


\paragraph{Human Detector and Testing}

We use a same human detector provided by the SimpleBaseline method~\cite{xiao_eccv_2018}\footnote{https://github.com/Microsoft/human-pose-estimation.pytorch}. It is based on Faster-RCNN~\cite{ren_nips_2015} with mAP of human category 56.4. During testing, Soft-NMS~\cite{bodla_iccv_2017} is used to suppress duplicated bounding boxes. As a common practice~\cite{chen_cvpr_2018,newell_eccv_2016,xiao_eccv_2018}, positions of keypoints are predicted on averaged heatmaps generated from original and flipped image.

\subsection{Analyses}

Several analyses are reported to show how FSM works in our model. Unless otherwise stated, all analyses are based on FSM3, which is the module inserted right before the first ESP \textit{E1}.

\begin{figure}[t]
  \centering
    \includegraphics[width=0.85\linewidth]{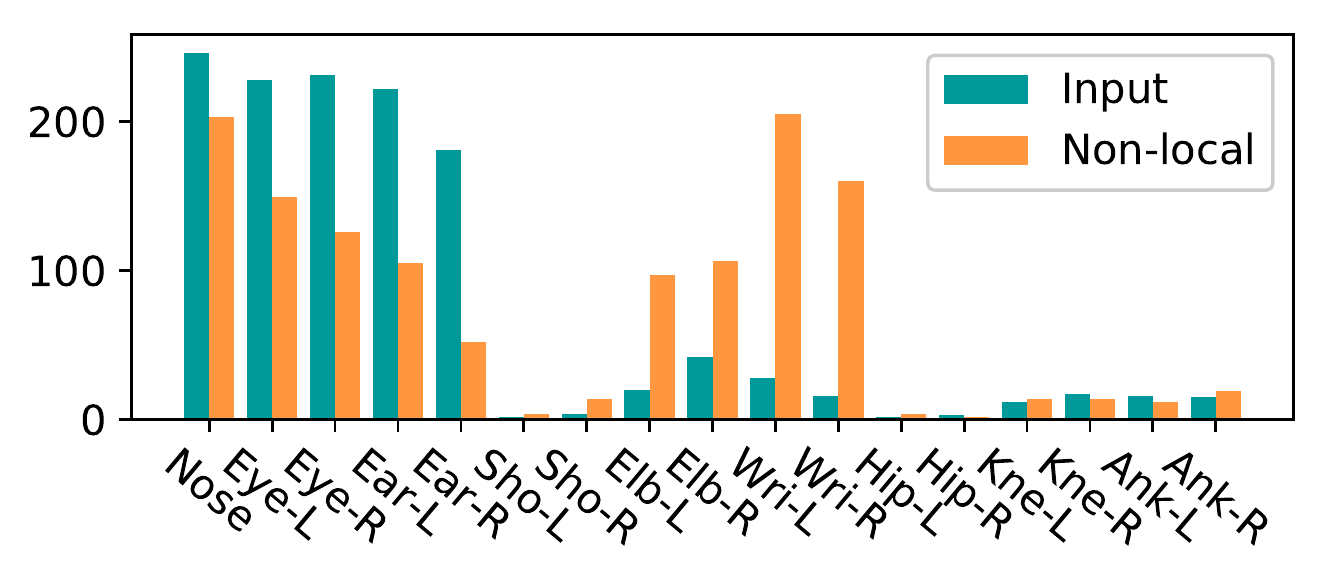}\\
    \includegraphics[width=0.85\linewidth]{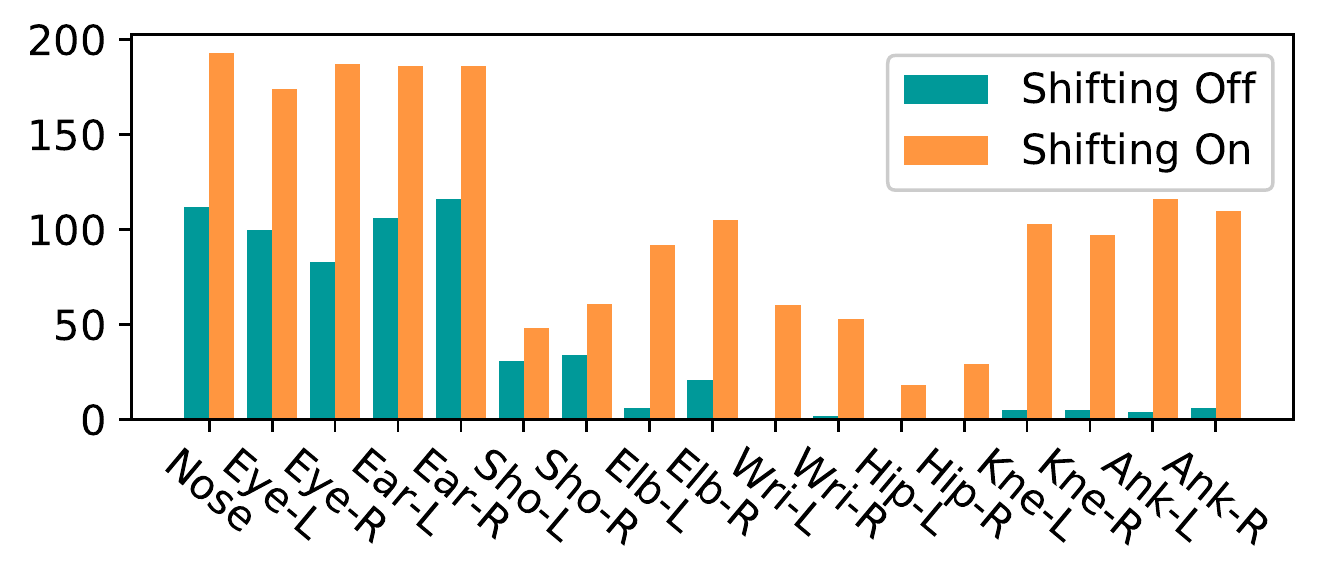}
  \caption{Comparison of channel contribution to keypoints. \textbf{Top}: Count of non-local and input channels with above-threshold contribution to keypoints. \textbf{Bottom}:Count of CA channels with above-threshold contribution to keypoints.}
  \label{fig:channel_contribution}
\end{figure}

\begin{figure}[t]
  \centering
  \includegraphics[width=0.8\linewidth]{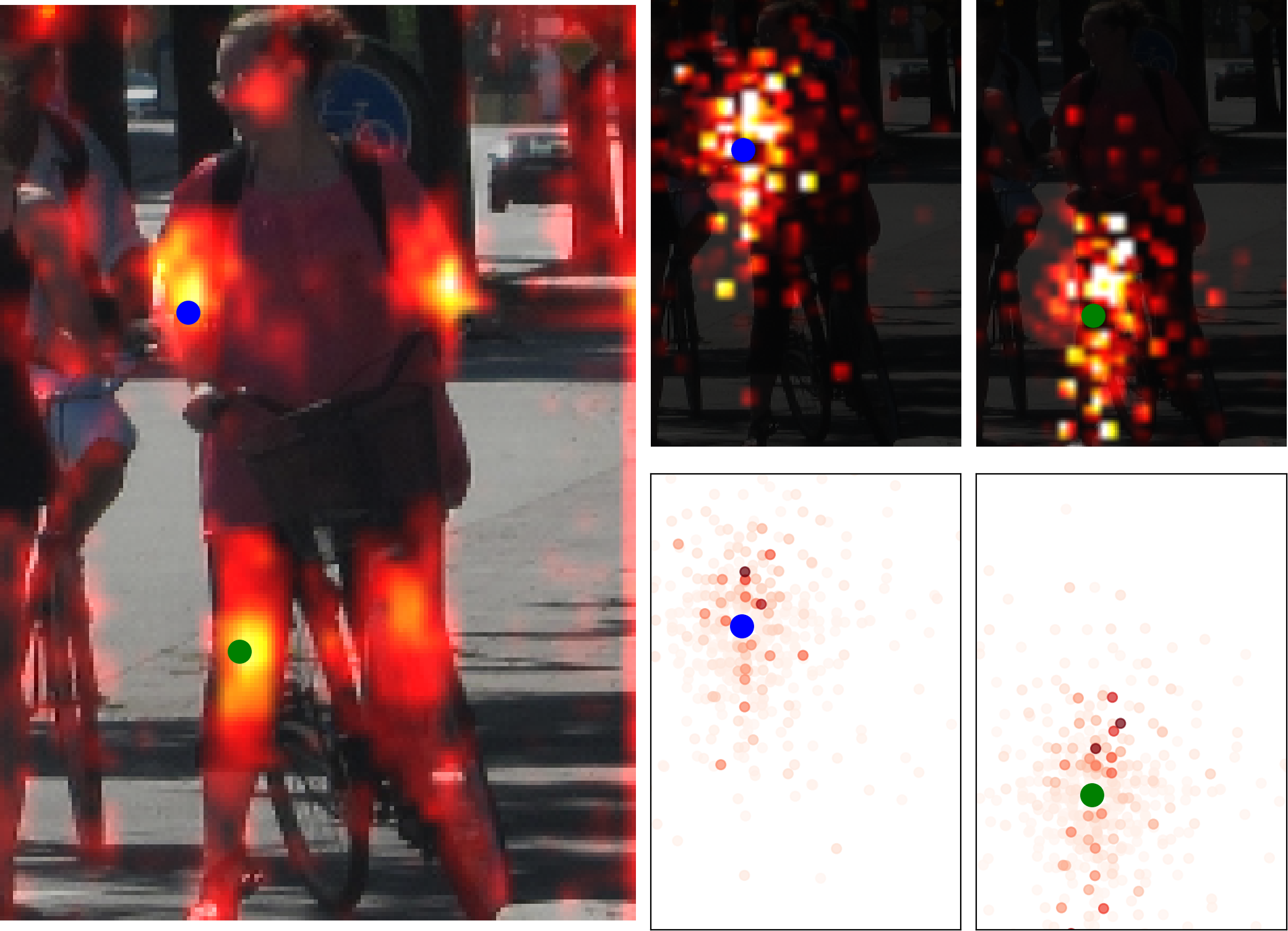}\\
  \includegraphics[width=0.8\linewidth]{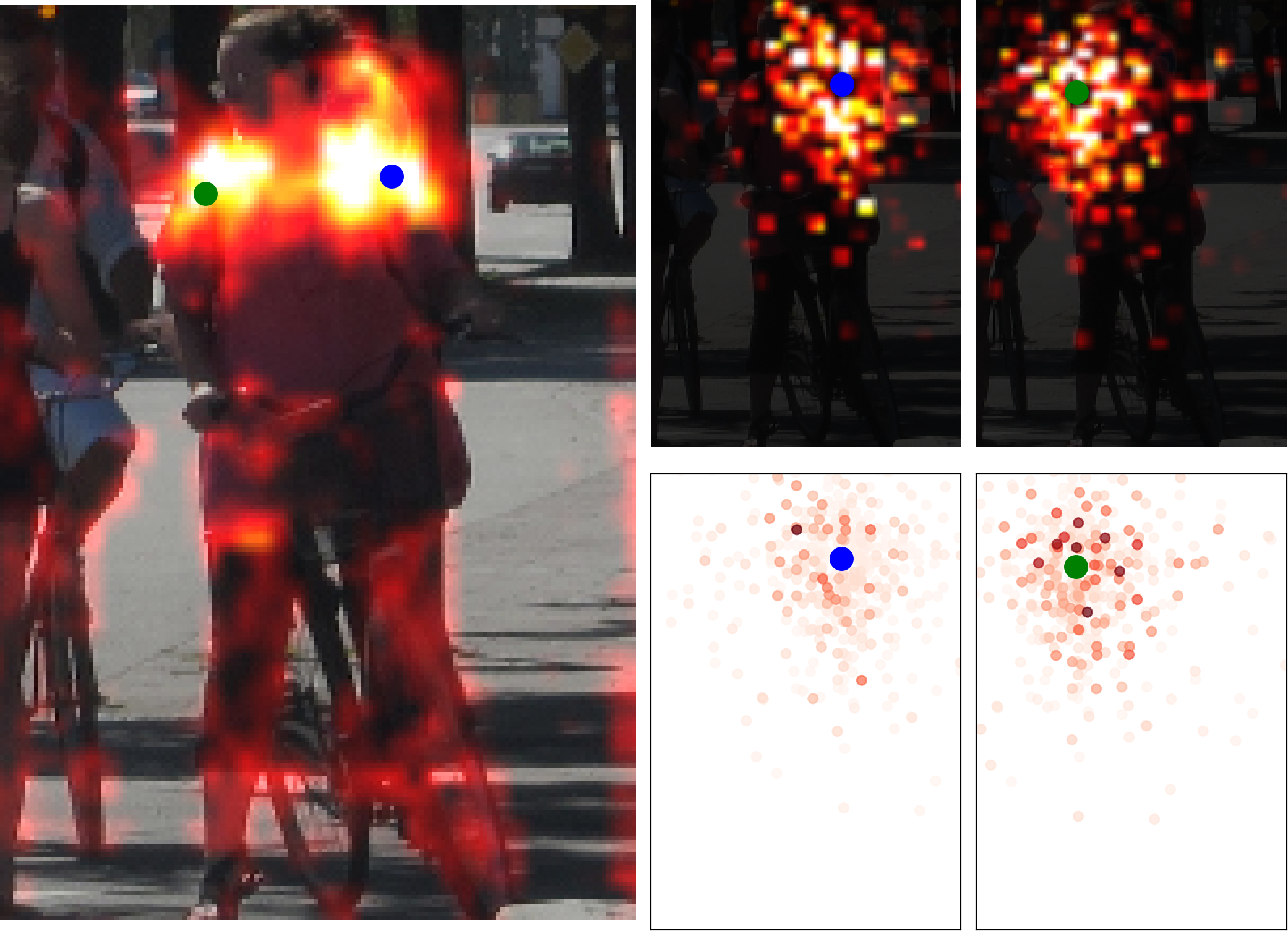}
  \caption{Non-local maps, effective receptive fields and window positions. \textbf{Big}: Non-local maps. \textbf{Small top}: Summed effective receptive fields on all input maps back-propagated respectively from blue and green points of a non-local map. \textbf{Small bottom}: Convolution window on input maps at blue and green points, with energy of \(\mathbf{w}^{\mathrm{FSM}}\) indicated by red color.}\label{fig:active_kernel}
\end{figure}

\paragraph{Learned Offsets}

Learned offsets of all FSMs except FSM3 are drawn in Figure~\ref{fig:all_offset}(a) and FSM3 is in Figure~\ref{fig:all_offset}(c). Distribution of these offsets shows some interesting pattern. First, most of the FSMs, especially deeper ones, have learned grid-aligned offsets. This phenomenon is probably caused by the correlation between pixels introduced from previous convolution and pooling layers. The grid effect become clearer at late stage of training, which suggest that small learning rate for offset at this time might induce offsets falling into local minima positions.
The second observation is, offsets are distributed around the center from densely to sparsely as the offsets grow larger. When FSM is viewed as convolution, this result agrees with RFB~\cite{liu_eccv_2018l} which enlarge effective receptive field by increasing density of kernels at surrounding area.

\paragraph{Contribution on Keypoint}

We especially show the offsets of FSM3, which is closed to the first ESP thus directly relevant to keypoint detection. We also visualize how the offsets contribute to each keypoint in Figure~\ref{fig:all_offset}(c). To obtain the relation between offsets and keypoints, we back-propagate from each keypoint prediction map to calculate scores between keypoint categories and shifting channels. The calculation of score is detailed in supplementary material.


The offsets with above-threshold scores are drawn. We choose 0.5 as the threshold to select most relevant offsets and in the same time to preserve enough number of offsets to demonstrate statistically significant results. Figure~\ref{fig:all_offset}(c) shows the selected offsets. Note that the concept of offset is for shifting maps, so the shape of convolution window formed by these offsets is origin-inversed. The result reveals that different window shapes are utilized for different keypoint categories. For example, shoulders' window focus less on upper pixels while more on lower pixels. The dependency of hips is vertically distributed. And detection of ankles, which is one of the most hard parts in pose estimation, significantly exploits surrounding information.


Denote the feature maps generated by the last \(1\times 1\) convolution layer of FSM as \textit{non-local maps}. We have also calculated scores between non-local channels and keypoint categories by the same method. The number of non-local channels with above-threshold scores for each keypoint is shown in the top of Figure~\ref{fig:channel_contribution}. For comparison, the count of input channels is also drawn in this figure. We can see that when the input channels mainly focus on detection of face keypoints, FSM3 transfer part of focus from face to upper limbs.

\paragraph{Spatial Dependency}

We visualize how the spatial dependency learned in FSM3 and demonstrate its ability in forming different convolution window depending on input in Figure~\ref{fig:active_kernel}. Two non-local maps with clear meaning are selected and drawn in the big sub-figure. To show effective receptive field, we back-propagate from two positions of each non-local map to input maps using the method described in \cite{luo_arxiv170104128_2017}. The obtained gradient maps are squared-summed over all input channels and shown in the small top sub-figures. We also draw window positions with the saturation of red color indicating energy of convolution weights \(\sum_{c'}{w_{c,k,c'}^{\mathrm{FSM}}}^2(\cdot) \) in the small bottom sub-figures.

From the visualized gradient and convolution window, we find that different subsets of window positions are utilized for different non-local channels. Furthermore, correlation attention filters the subset depending on what each position might be interested. For example, in the first non-local map, the window shape at the knee is wider and longer than that at the elbow. In the second non-local map, left and right shoulders both focus towards the human body instead of background. These observations are consistent across images, we will show more results in the supplementary material.

\begin{table}
  \centering
    \small
    \begin{tabular}{| l | c | c | c | c |}
      \hline
      Methods            & mAP           & \#Params & FLOPS \\
      \hline
      Backbone           & 69.7          & \hfill 25.1 M  &  \hfill 5.1 G  \\
      Backbone, w/ ESP   & 70.1          & \hfill 25.8 M  &  \hfill 5.9 G  \\
      SSN, w/o CA        & 71.3          & \hfill 35.6 M  &  \hfill 10.8 G  \\
      SSN, w/o shifting  & 71.2          & \hfill 40.5 M  &  \hfill 13.2 G  \\
      SSN, w/o ESP       & 72.0          & \hfill 39.8 M  &  \hfill 12.5 G  \\
      SSN, \(K=256\)     & 72.5          & \hfill 33.2 M  &  \hfill 9.6 G  \\
      SSN                & \textbf{73.0} & \hfill 40.5 M  &  \hfill 13.3 G \\
      \hline
      3Block + 3FSM, \(K=256\) & 62.9 & \hfill \textbf{0.8 M} & \hfill \textbf{2.5 G} \\
      3Block + 3FSM, \(K=512\) & 65.0 & \hfill 1.2 M & \hfill 3.9 G \\
      \hline
    \end{tabular}
  \caption{Performance on \textit{val2017}, parameter numbers and FLOPS. By default \(K=512\). For SSN-based methods, the first FSM has only halved \(K\). For 3Block + 3FSM methods, all FSMs have same \(K\).}\label{tab:ablation}
\end{table}

\begin{table*}
  \centering
    \resizebox{0.8\linewidth}{!}{%
    \small
    \begin{tabular}{| l | c@{\hspace{0.15cm}}p{1.5cm} c@{\hspace{0.15cm}}p{1.5cm} c@{\hspace{0.15cm}}p{1.5cm} c@{\hspace{0.15cm}}p{1.5cm} |}
      \hline
      \multirow{2}{*}{Methods} & \tikzmark{e1s}E1\tikzmark{e1e} & & \tikzmark{e2s}E2\tikzmark{e2e} & & \tikzmark{e3s}E3\tikzmark{e3e} & & & \\
      & \tikzmark{p4e}P4\tikzmark{p4s} & & \tikzmark{p3e}P3\tikzmark{p3s} & & \tikzmark{p2e}P2\tikzmark{p2s} & & \tikzmark{p1e}P1\tikzmark{p1s} & \\
      \hline
      \multirow{ 2}{*}{Backbone} & - & & - & & - & & & \\
      & 69.7 & & 69.6 & & 64.6 & & 23.0 & \\
      \hline
      \multirow{ 2}{*}{Backbone + FSM} & - & & - & & - & & & \\
      & 72.0 & & 71.9 & & 66.8 & & 24.1 & \\
      \hline
      \multirow{ 2}{*}{Backbone \qquad \quad + ESP} & 11.5 & & 47.7
      & & 63.2 & & & \\
      & 70.1 & (\(\uparrow_E 0.4\)) & 70.1 & (\(\uparrow_E 0.5\)) & 65.0 & (\(\uparrow_E 0.4\)) & 23.4 & (\(\uparrow_E 0.4\)) \\
      \hline
      \multirow{ 2}{*}{Backbone + FSM + ESP (SSN)} & 44.9 & (\(\uparrow_F \mathbf{33.4}\)) & 64.6 & (\(\uparrow_F \mathbf{16.9}\)) & 66.9 & (\(\uparrow_F \mathbf{3.7}\)) & & \\
      & 73.0 & (\(\uparrow_E \mathbf{1.0}\)) & 72.8 & (\(\uparrow_E \mathbf{0.9}\)) & 67.6 & (\(\uparrow_E \mathbf{0.8}\)) & 24.9 & (\(\uparrow_E \mathbf{0.8}\)) \\
      \hline
    \end{tabular}
    \begin{tikzpicture}[overlay, remember picture, shorten >=4pt, shorten <=4pt, line width=0.75pt]
    \draw [->] ([yshift=0.25\baselineskip]{pic cs:e1e}) -- ([yshift=0.25\baselineskip]{pic cs:e2s});
    \draw [->] ([yshift=0.25\baselineskip]{pic cs:e2e}) -- ([yshift=0.25\baselineskip]{pic cs:e3s});
    \draw [->] ([yshift=0.25\baselineskip]{pic cs:e3e}) .. controls +(1.7,0)  .. ([xshift=-5pt, yshift=0.7\baselineskip]{pic cs:p1s});
    \draw [->] ([yshift=0.25\baselineskip]{pic cs:p1e}) -- ([yshift=0.25\baselineskip]{pic cs:p2s});
    \draw [->] ([yshift=0.25\baselineskip]{pic cs:p2e}) -- ([yshift=0.25\baselineskip]{pic cs:p3s});
    \draw [->] ([yshift=0.25\baselineskip]{pic cs:p3e}) -- ([yshift=0.25\baselineskip]{pic cs:p4s});
    
    \draw [->,shorten <=2pt, shorten >=2pt] ([xshift=-2.5pt, yshift=0.25\baselineskip]{pic cs:e1s}) -- ([xshift=-2.5pt, yshift=0.25\baselineskip]{pic cs:p4e});
    \draw [->,shorten <=2pt, shorten >=2pt] ([xshift=-2.5pt, yshift=0.25\baselineskip]{pic cs:e2s}) -- ([xshift=-2.5pt, yshift=0.25\baselineskip]{pic cs:p3e});
    \draw [->,shorten <=2pt, shorten >=2pt] ([xshift=-2.5pt, yshift=0.25\baselineskip]{pic cs:e3s}) -- ([xshift=-2.5pt, yshift=0.25\baselineskip]{pic cs:p2e});
    \end{tikzpicture}
    }
  \caption{Performance in mAP of predictors on different stages evaluated on \textit{val2017}. The arrows in header row stands for network connections. E* are ESPs, P* are four original predictors in the backbone. \(\uparrow_E\) means the improvement made by introducing ESPs. \(\uparrow_F\) means the improvement made by introducing FSMs.}\label{tab:esp}
\end{table*}

\begin{table*}
  \small
  \centering
  \resizebox{\linewidth}{!}{%
    \begin{tabular}{| l | l c c c | c c c c c c |}
      \hline
      Methods & Backbone & Input Size & \#Params & FLOPS & AP & AP\textsubscript{@.5} & AP\textsubscript{@.75} & AP\textsubscript{m} & AP\textsubscript{l} & AR \\
      \hline
      
      CMU-Pose* \cite{cao_cvpr_2017} & - & - & - & - & 61.8 & 84.9 & 67.5 & 57.1 & 68.2 & 66.5 \\

      Mask-RCNN \cite{he_iccv_2017} & ResNet-50-FPN & - & - & - & 63.1 & 87.3 & 68.7 & 57.8 & 71.4 & - \\

      G-RMI \cite{papandreou_cvpr_2017} & ResNet-101 & \(353\times 257\) & 42.6 M & 57.0 G & 64.9 & 85.5 & 71.3 & 62.3 & 70.0 & 69.7 \\
      
      Associative Embedding* \cite{newell_nips_2017} & Hourglass & \(512\times 512\) & - & - & 65.5 & 86.8 & 72.3 &  60.6 & 72.6 & 70.2 \\
      
      Integral Pose Regression \cite{sun_eccv_2018} & ResNet-101 & \(256\times 256\) & 45.0 M & 11.0 G & 67.8 & 88.2 & 74.8 & 63.9 & 74.0 & - \\

      CPN \cite{chen_cvpr_2018} & ResNet-Inception & \(384\times 288\) & - & - & 72.1 & 91.4 & 80.0 & 68.7 & 77.2 & 78.5 \\
      
      RMPE \cite{fang_iccv_2017} & PyraNet \cite{xu_corr_2014} & \(320\times 256\) & 28.1 M & 26.7 G & 72.3 & 89.2 & 79.1 & 68.0 & 78.6 & - \\

      SimpleBaseline \cite{xiao_eccv_2018} & ResNet-101 & \(384\times 288\) & 53.0 M & 30.1 G & 73.2 & 91.4 & 80.9 & 69.7 & 79.5 & 78.6 \\
      
      SimpleBaseline \cite{xiao_eccv_2018} & ResNet-152 & \(384\times 288\) & 68.6 M & 35.6 G & \textbf{73.8} & \textbf{91.7} & \textbf{81.2} & \textbf{70.3} & \textbf{80.0} & 79.1 \\







      \hline

      SSN & ResNet-50-FPN & \(256\times 192\) & 40.5 M & 13.3 G & 72.4 & 91.4 & 80.2 & 69.0 & 77.4 & 78.7 \\
      SSN & ResNet-50-FPN & \(384\times 288\) & 40.5 M & 29.9 G & \textbf{73.7} & 91.6 & 80.7 & 70.1 & 78.9 & \textbf{79.7} \\
      
      \hline
      
      3Block + 3FSM, \(K=256\) & - & \(256\times 192\) & \textbf{0.8 M} & \textbf{2.5 G} & 62.5 & 86.8 & 68.3 & 59.5 & 66.9 & 68.9 \\
      3Block + 3FSM, \(K=512\) & - & \(256\times 192\) & 1.2 M & 3.9 G & 64.2 & 87.8 & 69.9 & 61.0 & 68.8 & 70.6 \\
      
      
      
      
      \hline
    \end{tabular}%
    }
  \caption{Results on COCO \textit{test-dev} dataset. * indicates bottom-up methods, the rest are top-down methods. Only the parameter number and FLOPS of body part detector in top-down methods are shown.}\label{tab:testdev}
\end{table*}

\subsection{Ablation Study}

\paragraph{Feature Shifting Module}

The contribution of each part is studied and reported in Table~\ref{tab:ablation}. The insertion of FSMs brings 2.9 mAP improvement from the backbone model also with ESP. We show that the improvement is not simply an additive result from each individual part of FSM, but a cooperative achievement.

Correlation attention plays a key role in the function of FSM. Without CA, the 9.8M parameters of FSM leads to only 1.2 mAP improvement, and the model is lower than standard SSN by 1.7 mAP. The distributions of all offsets and keypoint-related offsets trained in this configuration are shown in Figure~\ref{fig:all_offset}(b,d). It shows that smaller offsets have been learned, especially for FSMs at shallow layers.
For FSM3 without CA, less offsets are contributing to keypoint detection. The learned convolution window is also smaller especially for the detection of wrist, which is consistent to the observation from the top of Figure~\ref{fig:channel_contribution} that FSM3 is good at detecting wrist.

To prove that the contribution is also not solely brought by self-attention mechanism, we trained a model without feature map shifting and evaluated how the performance and the CA's contribution to keypoints changes. The performance without shifting largely drops by 1.8 mAP from SSN. The bottom sub-figure of Figure~\ref{fig:channel_contribution} shows the contribution of CA channels. Without shifting, the CA branch contribute less to keypoint detection, especially for lower body parts. This further provides evidence to the assumption that feature map shifting and attention mechanism have cooperatively improved the performance.

We also evaluate the result when \(K=256\) (again, exceptionally the first ESP has \(K=128\)). With halved shifting channels, the performance only drops by little.

To demonstrate FSM's ability in constructing lightweight networks, we build a very shallow network with only 3 Bottleneck blocks and 3 FSMs interleaved. Results in Table~\ref{tab:ablation} and Table~\ref{tab:testdev} shows they achieved competitive results with much smaller parameter number and computation cost. This shows FSM's high generalization ability and provides better choices for pose estimation on mobile devices.

\paragraph{Early Stage Predictors}

We compare under different configurations the performance made by all predictors, including the original ones and early stage predictors. Results are shown in Table~\ref{tab:esp}. The results shows that without FSMs, the performance made at ESPs are bad, especially the first ESP has only 11.5 mAP. However, after introducing FSMs, all ESPs have been largely improved, and the first ESP has made 33.4 mAP increment. The introducing of ESP has also made the final performance increase by 1 mAP. This result proves FSMs can boost the detection ability at early stage, with the help of ESPs.


\subsection{Results on COCO keypoint test-dev}

The architecture and performance comparison between our SSN and other methods are reported in Table~\ref{tab:testdev}. Our SSN model outperform many other larger methods, showing its ability in achieving better performance by modeling long-range dependency. The 3Block+3FSM models with much fewer parameters have made very promising results, and even also outperform some large top-down methods.

\section{Conclusion}

In this paper we propose the spatial shortcut network which integrate feature shifting modules and correlation attention mechanism. The module improve shallow layers' detection ablility by modeling long-range spatial dependency. We demonstrated how the proposed module contribute to pose estimation task and present SSN's better performance with similar or smaller architecture. The spatial dependency can be further improved by regressing or guiding on the offsets, which we will explore in the future.

  {\small
    \bibliographystyle{ieee}
    \bibliography{egbib}
  }

\end{document}


\title{Supplementary Materials for\\\textit{Spatial Shortcut Network for Human Pose Estimation}}


\maketitle

\vspace*{-4em}
\thispagestyle{empty}


\section{Keypoint-offset score}

To calculate the scores between keypoint categories and shifting channels (or offsets), we use a back-propagation method, obtaining the partial gradient of keypoint prediction with respect to post-shifting maps. For each keypoint category \(m\), we firstly generate pseudo ground truth maps, which are copies of prediction maps produced by the model and have \(M\) channels. Then at the maximum position of the \(m\)-th channel of generated maps we modify the value to zero. Mean squared error loss between the modified maps and prediction maps are calculated and back-propagated from, and gradients with respect to all \(K\) post-shifting maps are generated. Finally, the gradients are averaged spatially, producing the scores between \(m\)-th keypoint and all \(K\) shifting channels. After loop through all \(M\) keypoint categories, we normalize the \(M\times K\) scores within each shifting channel.

\section{More spatial dependency visualization}

\begin{figure*}
    \centering
    \includegraphics[width=\linewidth]{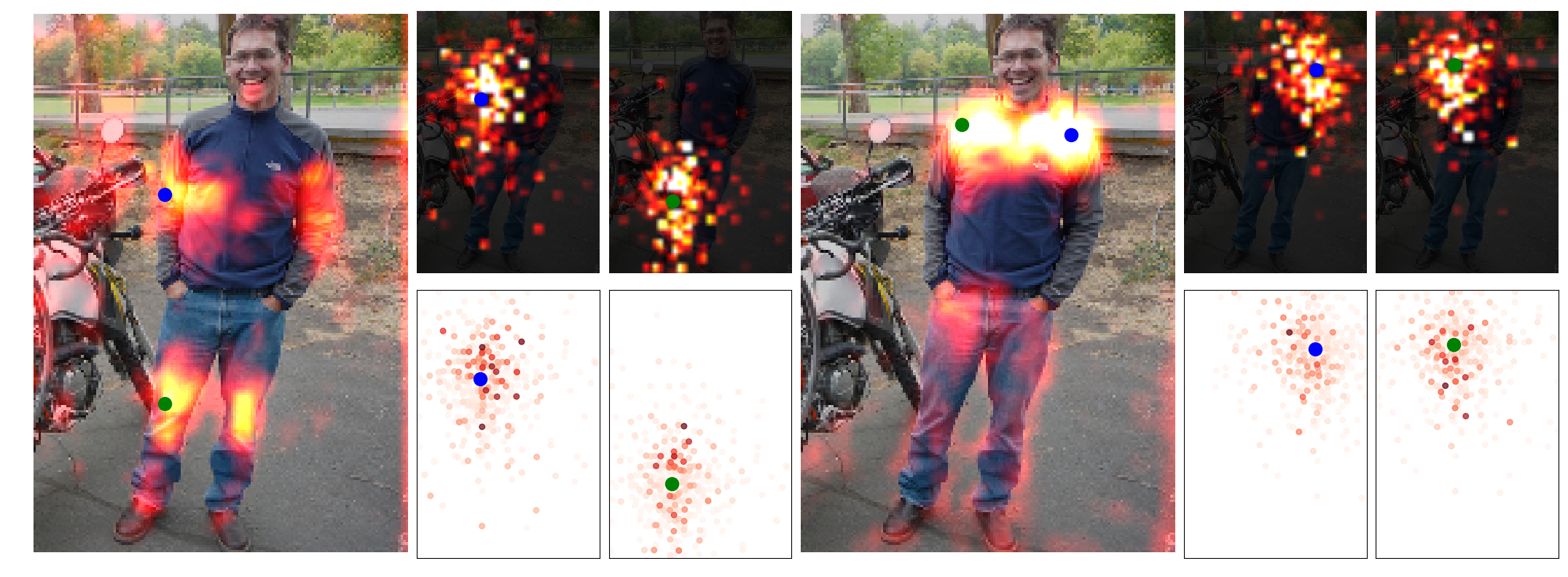}
    \includegraphics[width=\linewidth]{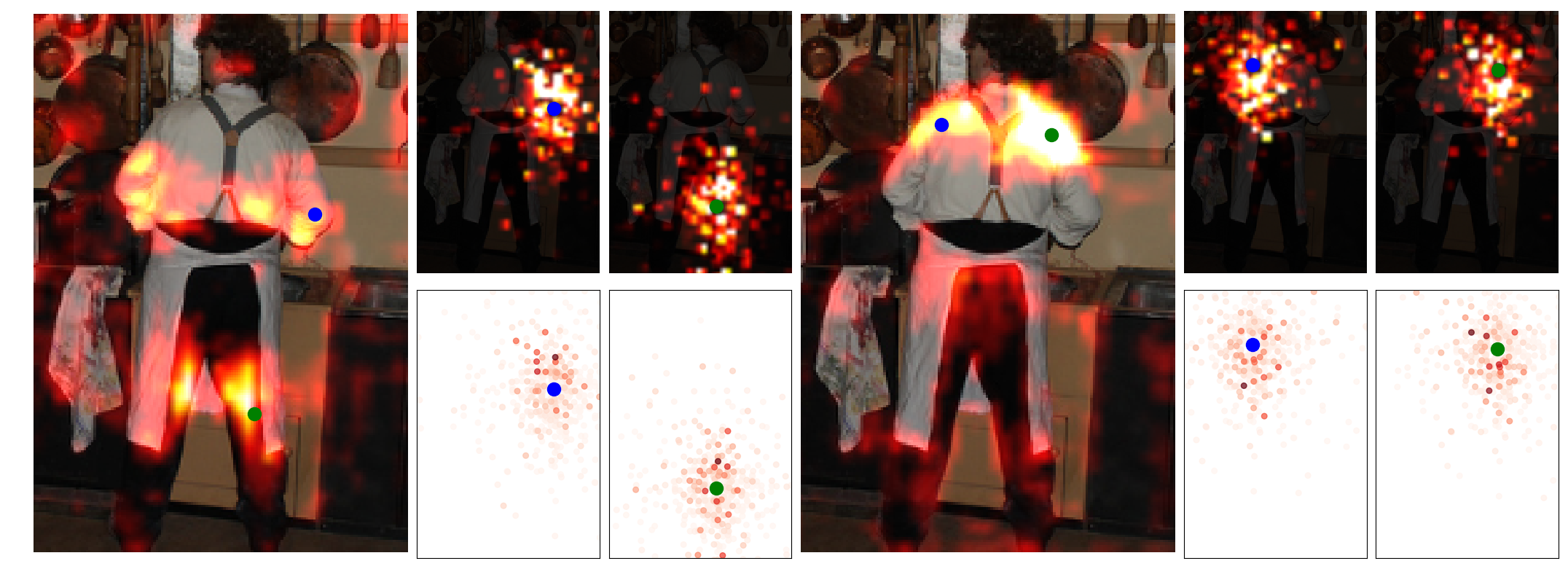}
    \includegraphics[width=\linewidth]{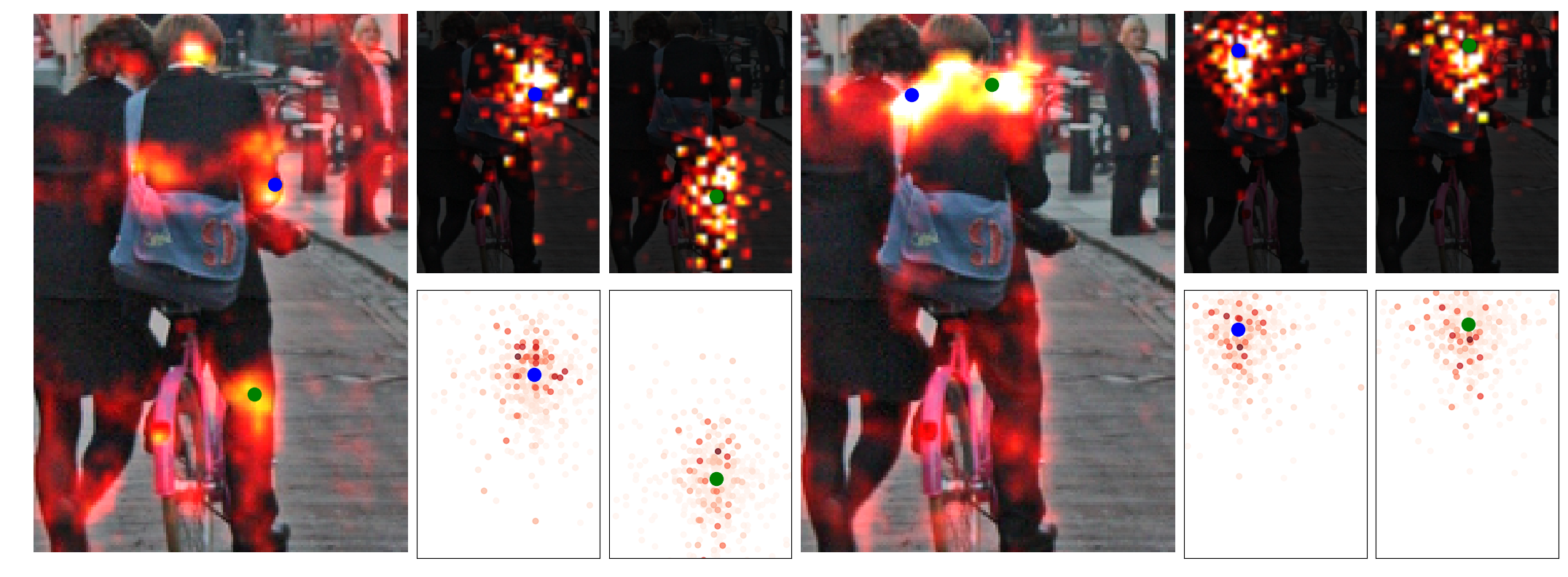}
    \caption{More results of non-local maps, effective receptive fields and window positions. \textbf{Big}: Non-local maps. \textbf{Small top}: Summed effective receptive fields on all input maps back-propagated respectively from blue and green points of a non-local map. \textbf{Small bottom}: Convolution window on input maps at blue and green points, with energy of \(\mathbf{w}^{\mathrm{FSM}}\) indicated by red color.}
    \label{fig:spatial_dep}
\end{figure*}

We show more spatial dependency for different images in Figure~\ref{fig:spatial_dep}. The results show more evidence about that the formed window shapes can be different depending on input data.

\section{Details of 3Block+3FSM network}

The structure of the 3Block+3FSM network is shown in Table~\ref{tab:structure}.

For FSM, we use a slightly different CA branch in this network. The original CA branch has spatial normalization at last, which encourage FSM to focus on most important positions. This focusing can make the learned spatial dependency specialized for particular kind of body part. The main model SSN converges faster and better with spatial normalization. However in this network, we found by visualization that for shallow layers the produced focusing was not very stable, causing the network to perform badly. So we removed the spatial normalization layer, and also replaced the Softplus function with Sigmoid function to keep the attention value between 0 and 1.

\begin{table}[ht]
\centering
\begin{tabular}{|l|c|c|c|}
\hline
Type & Patch size/stride & Output size \\
\hline\hline
input & & \(3 \times 256 \times 192\) \\
\hline
conv-GN-ReLU & \(7\times 7 / 2\) & \(64\times 128 \times 96\) \\
\hline
max pool & \(3\times 3 / 2\) & \(64\times 64 \times 48\) \\
\hline
FSM &  & \(64\times 64 \times 48\) \\
\hline
Bottleneck &  & \(256\times 64 \times 48\) \\
\hline
FSM &  & \(256\times 64 \times 48\) \\
\hline
Bottleneck &  & \(256\times 64 \times 48\) \\
\hline
FSM &  & \(256\times 64 \times 48\) \\
\hline
Bottleneck &  & \(256\times 64 \times 48\) \\
\hline
conv-GN-ReLU & \(1\times 1 / 1\) & \(256\times 64 \times 48\) \\
\hline
conv-BN & \(3\times 3 / 1\) & \(17\times 64 \times 48\) \\
\hline

\end{tabular}

\caption{Structure of 3Block+3FSM. GN and BN stand for group normalization and batch normalization respectively.}\label{tab:structure}
\end{table}
